\documentclass[10pt,twocolumn,letterpaper]{article}
 
\usepackage{cvpr}
\usepackage{times}
\usepackage{epsfig}
\usepackage{graphicx}
\usepackage{amsmath}
\usepackage{amssymb}
\usepackage{amsthm}
\usepackage{placeins}
\usepackage{dsfont} 
\usepackage{wrapfig}

\newcommand{\ppi}{p\oplus i}
\newcommand{\pmi}{p\ominus i}
\newcommand{\suppress}[1]{}

\newtheorem{theorem}{Theorem}

\usepackage[pagebackref=true,breaklinks=true,letterpaper=true,colorlinks,bookmarks=false]{hyperref}

\cvprfinalcopy % *** Uncomment this line for the final submission 

\def\cvprPaperID{1567} % *** Enter the CVPR Paper ID here

\ifcvprfinal\fi
\begin{document}

%%%%%%%%% TITLE

\title{Efficient Regularization of Squared Curvature} 
%\title{Efficient Squared Curvature Model} 
%\title{Efficient Squared Curvature Optimization} 
%\title{Fast Squared Curvature} 
%\title{Fast Squared Curvature Regularization}  
%\title{Fast Squared Curvature Regularization via LSA-TR}  

\author{ 
Claudia  Nieuwenhuis  \hspace{3ex}     Eno  Toeppe      \hspace{3ex}    
Lena  Gorelick   \hspace{3ex}         Olga  Veksler    \hspace{3ex}         Yuri  Boykov    \\[1ex]
Computer Science Department \\
University of Western Ontario  
% \and
% \and
}

\maketitle
%\thispagestyle{empty}

%%%%%%%%% ABSTRACT
\begin{abstract}
  Curvature has received increased attention as an important
  alternative to length based regularization in computer vision. In
  contrast to length, it preserves elongated structures and fine
  details. Existing approaches are either inefficient, or have low angular
  resolution and yield results with strong block artifacts.
%  Existing approaches are inefficient, have low angular
%  resolution and yield results with strong block artifacts. 
We derive
  a new model for computing squared curvature based on integral
  geometry. The model counts responses of straight line triple
  cliques. The corresponding energy decomposes into
  submodular and supermodular pairwise potentials.
  We show that this energy can be efficiently minimized even for high
  angular resolutions using the \emph{trust region} framework. Our results confirm that we
 % outperform existing methods with respect to quality and runtime.
 obtain accurate and visually pleasing solutions without strong artifacts at reasonable runtimes.

%We define a novel regularizer and show how to efficiently minimize the corresponding energy for high angular resolution. 

\end{abstract}

%%%%%%%%% BODY TEXT

\section{Introduction}
\begin{figure}
\centering
\tabcolsep0.5mm 
\begin{tabular}{cccc}
\includegraphics[width = 0.4\linewidth]{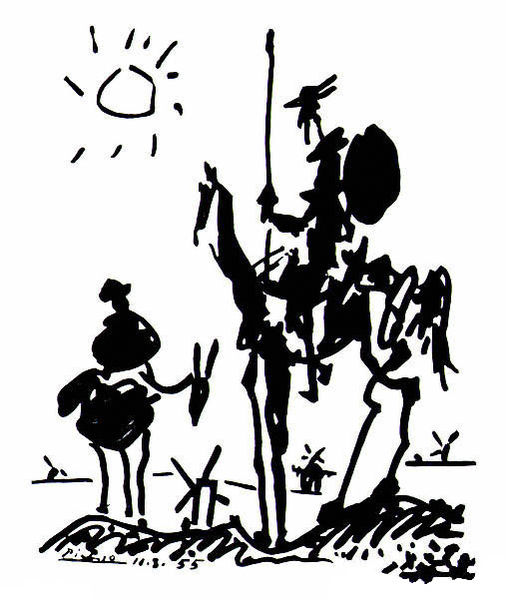}&
\includegraphics[width = 0.4\linewidth]{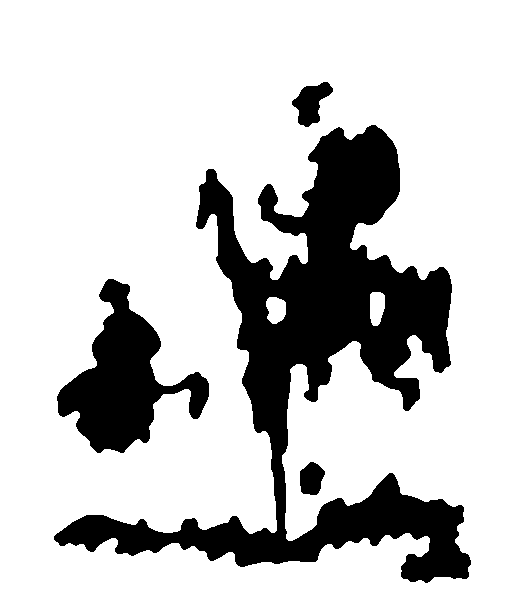} \\
a) Original & (b) Length Regularization \\
\includegraphics[width = 0.4\linewidth]{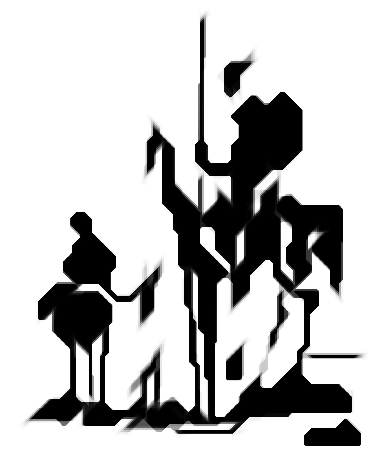} &
\includegraphics[width = 0.4\linewidth]{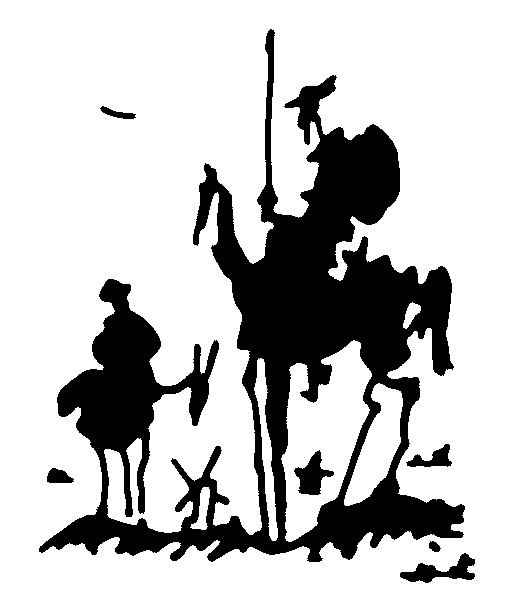} \\
 (c) Elastica Regularization \cite{heber-et-al-eccv-2012} & (d) Proposed 
\end{tabular}   
\caption{Segmentation results of Picasso's ink drawing using (b) length based regularization computed with the Potts model based on a 16 neighborhood using \cite{boykov-jolly-iccv-2001}, (c) Euler's elastica consisting of squared curvature and length regularization computed by Heber et al. \cite{heber-et-al-eccv-2012}, (d) the novel squared curvature regularizer.}
\label{fig:teaser}
\end{figure}

A number of vision tasks can be formulated as an energy minimization
problem such as segmentation, 3D reconstruction, stereo and
inpainting. The corresponding energies typically consist of a data
affiliation term and a regularization term. The data affiliation term
relates the solution to the image data, while the regularization term
imposes some kind of prior knowledge on the result. Length-based
regularizers give rise to sub-modular energies (Potts model) that can
be globally and efficiently optimized and, therefore, are widely
used. Their main disadvantage is an inherent shrinking bias, which
tends to eliminate thin and elongated structures such as vessels or
limbs, see Figure \ref{fig:teaser}(b).  In contrast, curvature-based
regularizers preserve such fine details, but are more difficult to
model, often resulting in non-submodular energies that are hard to
optimize.

Previous approaches to curvature regularization are computationally intensive
\cite{schoenemann-etal-ijcv-2012,strandmark-kahl-emmcvpr-2011}, have strong discretization artifacts \cite{schoenemann-etal-ijcv-2012,strandmark-kahl-emmcvpr-2011,elzehiry-grady-cvpr-2010,heber-et-al-eccv-2012,shekhovtsov-etal-dagm-2012}
and are often restricted to specific angular resolutions
\cite{elzehiry-grady-cvpr-2010} or grid complexes
\cite{schoenemann-etal-ijcv-2012,strandmark-kahl-emmcvpr-2011,heber-et-al-eccv-2012}. Curvature is often combined with length to compute Euler's elastica, which can alleviate some of the problems of pure curvature regularizers, see Figure \ref{fig:teaser}(c). However, satisfactory solutions to the curvature regularization problem have not been found yet. 

In this paper we present a novel approach to modeling and efficiently 
optimizing squared curvature, which yields visually pleasing results without strong discretization artifacts, see Figure \ref{fig:teaser}(d).

Our goal is to regularize the boundary of a binary labeling $S$ based on an integral of squared curvature
\begin{equation}\label{eq:contK}
K(S)= \int_{\partial S} \kappa^2 \cdot ds.
\end{equation} 
For example, a segmentation energy can combine this regularization term with a
regional appearance term $\int_{int(S)} f(p)\, dp$ for some potential function $f$ over pixels $p$. 

We optimize a binary energy function with triple clique potentials, which can be reduced to an integer quadratic energy with both submodular and supermodular pairwise terms without adding auxiliary variables.

 Such energies can be efficiently optimized with the LSA-TR method proposed in our companion paper \cite{LSATR:companion}. Our formalism is based on integral geometry, allows for high angular resolutions and yields excellent results compared to previous approaches.

Our contributions can be summarized as follows \vspace{-0.2cm}
\begin{itemize}
\item We propose a novel model for measuring squared curvature based
  on integral geometry and show how it relates to counting straight triple cliques. \vspace{-0.2cm}
\item Our model can be formulated as a pairwise quadratic energy and optimized efficiently even for high angular resolutions. \vspace{-0.2cm}
\item The proposed approach outperforms previous methods in terms of quality of the results as well as efficiency. \vspace{-0.2cm}
\end{itemize}

{\bf Related work:} Most existing models for curvature regularization in vision  are based on the Bruckstein formula \cite{schoenemann-etal-ijcv-2012, strandmark-kahl-emmcvpr-2011, heber-et-al-eccv-2012, elzehiry-grady-cvpr-2010} except for \cite{shekhovtsov-etal-dagm-2012} which is based on learning.

Shekhovtsov et al. \cite{shekhovtsov-etal-dagm-2012} learn the costs for curvature in the form of ``soft'' patterns which serve as filters whose response is locally minimized. However, their MRF approach lacks accuracy due to missing consistency constraints between neighboring patterns.

Approaches based on the Bruckstein formula \cite{Bruckstein01} express curvature as the exterior angle sum of an approximating polygon. Mostly this is done by formulating an optimization problem on a cell complex, which is a planar graph with fixed, regular structure. A segmentation is a consistent subset of faces and edges. Local ``curvature'' is then measured by the exterior angle between consecutive boundary edges.

The pioneering approach to image segmentation with curvature regularization on cell complexes was given by Schoenemann et al. \cite{schoenemann-etal-ijcv-2012}. They solve a linear program where each variable corresponds to a configuration of two edges with associated angle. 
%Consistency of active edges and regions are enforced by constraints. The segmentation problem can, thus, be formulated as an integer linear %programming that is relaxed and optimized independent of initialization. Curvature is approximated locally by penalizing the angle between adjacent %active boundary edges.
Strandmark and Kahl \cite{strandmark-kahl-emmcvpr-2011} improve this framework by removing extraneous arcs and generalize it to 3D surfaces. 
%They also propose a generalization to 3D segmentation and give an analysis of different cell structures. 
Heber et al. \cite{heber-et-al-eccv-2012} provide a formulation of the curvature model on cell complexes that can be optimized by approximating the envelope of the underlying non-convex functional. 

The main drawback of the approaches formulated on cell complexes is their high runtime up to several hours, which is due to the large number of variables and consistency constraints. In addition, these methods suffer from a strong angular bias since they only allow for specific edge configurations. Although angular resolution can in theory be increased arbitrarily, solving the problem for high resolutions easily becomes infeasible. This is not the case for our approach since the number of triple cliques grows linearly with the angular resolution. 

Another approach based on the Bruckstein formula which is related to cell complexes is El-Zehiry and Grady's work \cite{elzehiry-grady-cvpr-2010}. They formulate their problem on a regular pixel grid, which is interpreted as a cell complex. Accordingly, their angular resolution is limited by 90 degrees, which leads to a coarse approximation of curvature. Extending this approach to a higher angular resolution is not possible due to  inconsistencies with the Bruckstein formula.

Our method differs from the previous approaches in that we neither explicitly model our segmentation boundary as a polygon by means of cell complexes, nor do we measure angles between edges. Instead, we count the number of ``active'' straight line triple cliques and relate it to squared curvature based on integral geometry.

%All the listed approaches measure curvature based on the Bruckstein formular \cite{bruckstein}. Our model is not based on this formula, instead we %measure the number of "active" three-cliques forming straight lines, which can theoretically be related to measuring curvature.

%Schoenemann, Strandmark, Grady, Pock, Rother
%Unlike Grady, all other methods (including ours) can measure curvature more accurately 
%than at 90 degree angles.
%
%- all approaches are based on bruckstein formula
%- Schoenemann: * first approach 
%							 * defined on a complex and formalized as relaxed LP
%							 * problems: slow and artifacts because of complexes
%							 
%- Grady: * measuring turns of pixel boundaries
%				 * also based on triple cliques with very similar enery but very different model
%				 * optimized with QPBO
%				 * problems: only for 90 degree turns and not well optimizable for weak data term

\section{Integrating Squared Curvature}

We propose a new discrete model for approximating the squared curvature integral $K(S)$ in \eqref{eq:contK}
based on a certain class of triple cliques. Our combinatorial approach is presented below in the context of 2D segmentation and inpainting. 
It has straightforward extensions to segmentation problems in higher dimensions, but this is left for future work, 
see Section~\ref{sec:conclusions}. Unlike previous discrete
\begin{wrapfigure}{r}{0.42\linewidth}
 \centering
 \vspace{-9pt}
 \hspace{-15pt} 
\includegraphics[width=1.1\linewidth]{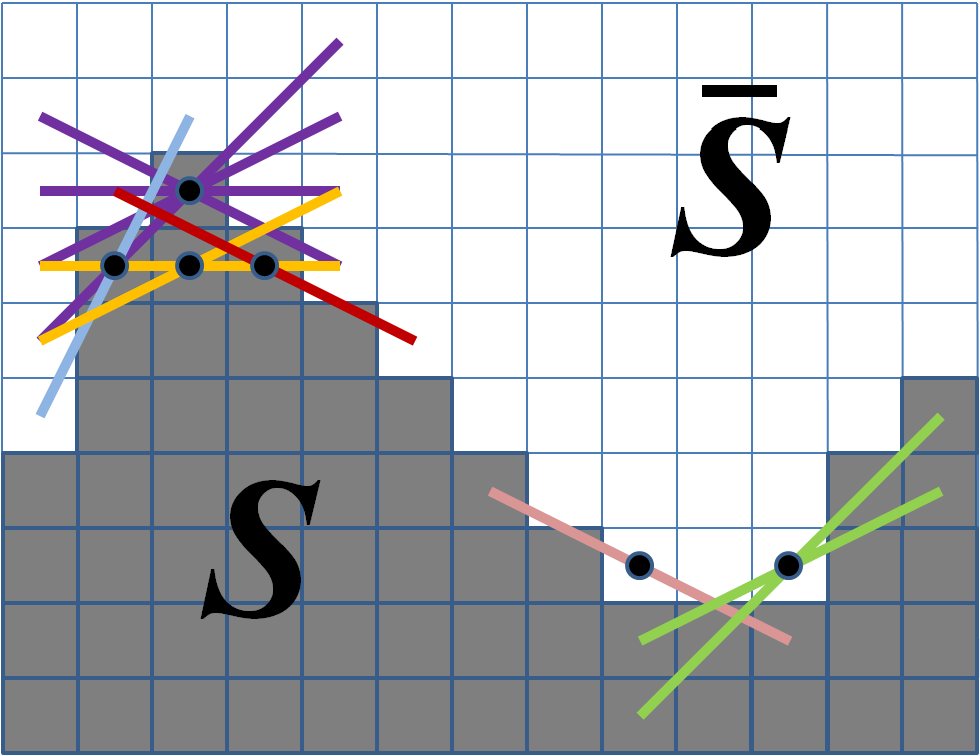}
 \vspace{-9pt}
\end{wrapfigure}
methods for curvature, we use a neighborhood system of symmetric 'straight' triple cliques.
Our  general intuition is illustrated on the right. Let the indicator variables $x_p$ on pixels $p\in\Omega$ 
define the segment $S=\{p\,|\,x_p=1\}\subset\Omega$ and consider a collection of triple cliques  $(p-\Delta,p,p+\Delta)$ for each pixel $p$ and 
its symmetric neighbors for fixed 2D shifts $\pm \Delta$ within a $5\times5$ neighborhood. 
The colors show triple cliques with binary configurations $(0,1,0)$ or $(1,0,1)$
at different pixels $p$. Such triple cliques 'fire'  only on curved parts of the boundary of $S$.
The number of responses increases for larger curvature. Configurations $(0,1,0)$ respond to positive and $(1,0,1)$ 
to negative curvature. These observations suggest that such triple cliques can measure curvature.
The details are presented below.

\subsection{Notation: Variables, Cliques, Neighborhood}
The proposed neighborhood structure is somewhat similar to the standard regular neighborhood 
of pairwise cliques (edges) commonly used for length-based regularization \cite{BK:iccv03}.
At each pixel $p$ the neighborhood is bounded, \eg by a $5\times5$ box centered at this pixel, see Figure \ref{fig:blob1}(b).
Interactions are imposed between pixel $p$ and its $m$ neighbors in a certain discrete set of
directions $\Theta = \{\theta_i\,|\,i=1,..,m\}$ limited by the grid locations within the box. 
Standard length-based regularization \cite{BK:iccv03} corresponds to pairwise Potts interactions imposed 
over $p$ and each neighbor $q=\ppi:=p+\bar{d}_i$ where $\bar{d}_i$ denotes a shift of given length $d_i$ 
in direction $\theta_i$. Our new curvature-based regularization model uses interaction potentials imposed over
$p$ and pairs of symmetric neighbors in directions $\bar{d}_i$ and  $-\bar{d}_i$ forming a straight triple clique 
\begin{equation} \label{eq:tripleclique}
c_{i}(p) := (\pmi,p,\ppi)
\end{equation}
centered at pixel $p$, see Figure \ref{fig:blob1}(b).
We will use standard binary variables $x_p\in\{0,1\}$ to denote the object/background label at pixel $p$.
Vector $X=(x_p\,|\,p\in\Omega)$ will denote a configuration of binary labels of all pixels in $\Omega$
defining segment $S=\{p\,|\,x_p=1\}\subset\Omega$. 
Vector $X_c=(x_{\pmi},x_p,x_{\ppi})$ will denote a configuration of labels for pixels in clique $c_i(p)$. 

Each triple clique $c_i$ centered at pixel $p$ is described by its orientation $\theta_i$ and distance $d_i$, 
see Figure \ref{fig:blob1}(b). We will also use $\Delta\theta_i := \theta_{i+1}-\theta_i$ to denote the angular rotation 
to the next clique. Since our triple cliques include symmetric pairs of neighbors, the actual number of distinct 
triple cliques $c_i(p)$ centered at pixel $p$ is half the number of its neighbors. To avoid confusion, 
in the rest of the paper $m$ denotes the number of distinct triple cliques $c_i$ at pixel $p$ where $i\in\{1,\dots,m\}$ 
is an index of orientation in the set $\Theta = \{\theta_i\,|\,i=1,..,m\}$ such that $\Theta\subset [0,\pi)$.
This avoids identical cliques with orientations $\theta$ and $\pi+\theta$.

\subsection{Partial Sum Approximation of \eqref{eq:contK}}

\begin{figure}
\begin{tabular}{cc}
\includegraphics[width = 0.42\linewidth]{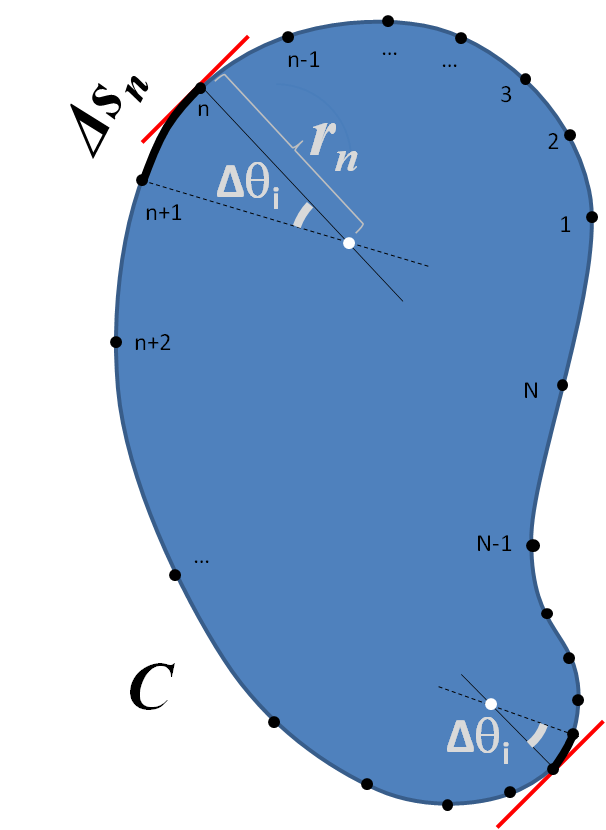} &
\includegraphics[width = 0.5\linewidth]{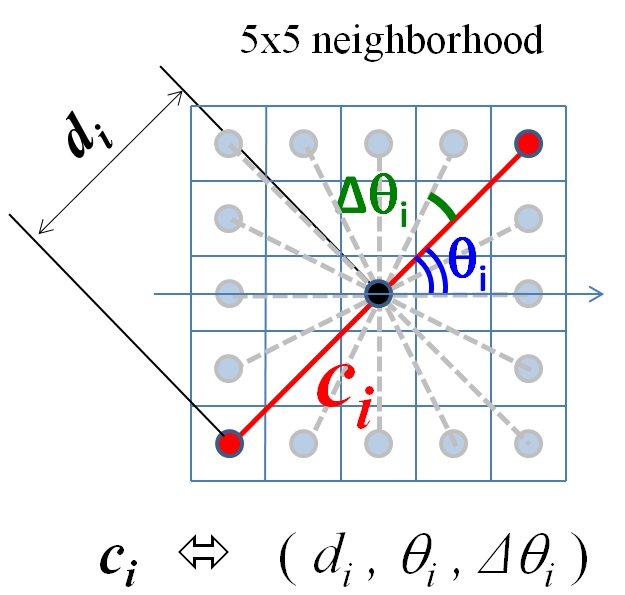}\\
(a) Contour points $\{n\}$ & (b) 3-clique of orientation $\theta_i$
\end{tabular}
\caption{The partial sum approximation \eqref{eq:contK_partial} of the
  integral $ \int_C \kappa^2 ds$ uses a sequence of points on the contour
  $C$. Our approach (a) is to select all contour points
  $\{n\}_{1}^{N}$ where the tangent is consistent with the orientation of some
  triple clique in a given neighborhood (b). For example, (a) marks
  two points with a tangent of the same orientation as clique $c_i$ in
  (b). Each point $n$ corresponds to a particular clique index
  $i(n)$. }
\label{fig:blob1}
\end{figure}

\begin{figure}
\begin{tabular}{c}
\includegraphics[width = 0.95\linewidth]{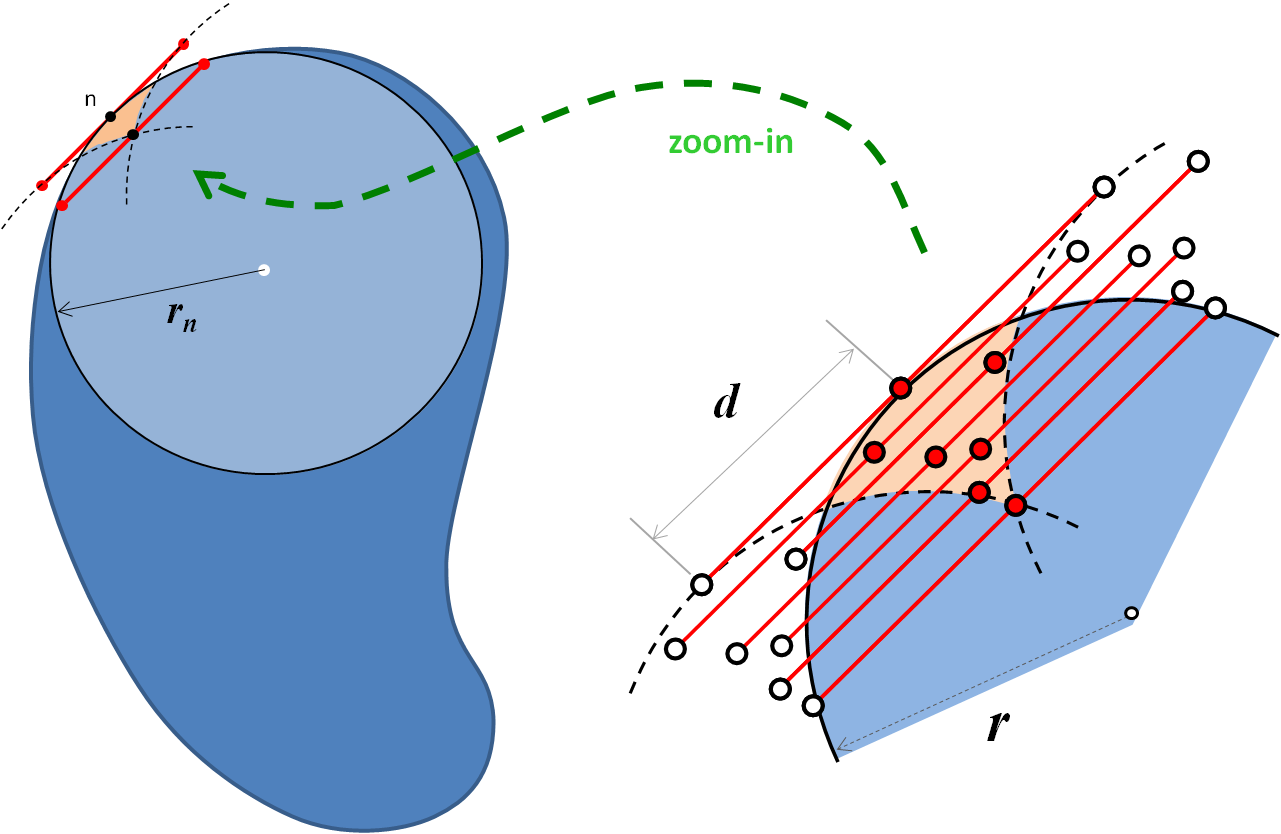}
\end{tabular}
\caption{Examples of triple cliques of orientation $\theta_i$ that ``fire'' in a vicinity of point $n$ on contour $C$ in Figure \ref{fig:blob1}(a). 
The brown region highlights points $p$ where clique $c_i(p)= (\pmi,p,\ppi)$ has configuration $X_c=(0,1,0)$.
The area of this region $A(\kappa,d)$ in \eqref{eq:taylor}, that is the number of such triple cliques, is determined by local curvature 
$\kappa_n=\frac{1}{r_n}$ and by clique size parameter $d_i$. }
\label{fig:blob2}
\end{figure}

Our discrete model for evaluating the integral of squared curvature \eqref{eq:contK}
is based on a partial sum approximation. Without loss of generality, assume that
the (continuous) segment $S$ has genus zero and that its boundary $\partial S$ 
is a closed contour $C$, as shown in Figure \ref{fig:blob1}(a). Then, the integral in \eqref{eq:contK}
can be approximated as
\begin{equation}\label{eq:contK_partial}
\int_C \kappa^2 \cdot ds \;\;\approx\;\; \sum_{n=1}^N \kappa^2_n \cdot \Delta s_n
\end{equation}
where $\{n\}_{1}^{N}$ is a finite sequence of points $n\in C$, $\Delta s_n$ is a contour length 
between adjacent points, and $\kappa_n$ is the curvature at point $n$.
In general, such approximations converge for finer discretizations $\{n\}_{1}^{N}$ 
as $max_n |\Delta s_n|\rightarrow 0$ if the contour $C$ is sufficiently smooth.

It is common to select a sequence of approximating points $\{n\}_{1}^{N}$ at equal intervals.
Our approach selects these points differently. We choose
points $\{n\}_{1}^{N}$ where tangents coincide with orientations
$\Theta = \{\theta_i\,|\,i=1,..,m\}$ for a chosen neighborhood system, 
as illustrated in Figure \ref{fig:blob1}(a,b). Assume that curvature is nearly constant between adjacent
points, then $\Delta s_n \approx r_n\cdot\Delta\theta_{i(n)} = \frac{\Delta\theta_{i(n)}}{|\kappa_n|}$ where $r_n$ is
the radius of the osculating circle at point $n$ and $\theta_{i(n)} \in\Theta$ is the tangent orientation at point $n$.
Thus, the partial sum \eqref{eq:contK_partial}  becomes
\begin{equation}\label{eq:contK_approx}
 \sum_{n=1}^N \kappa^2_n \cdot \Delta s_n 
 \;\approx\; \sum_{n=1}^N |\kappa_n| \cdot \Delta\theta_{i(n)}.
\end{equation}

\begin{theorem} \label{th:area}
Let contour point $n$ have tangent orientation $\bar{u}$ and osculating ball $B$ of radius 
$r=\frac{1}{|\kappa|}$. Then, the set of all points $p\in B$ such that $||p-n||\leq r$ 
and $(p\pm d\cdot\bar{u})\not\in B$ for given distance $d<r$ 
has area $A(\kappa,d)=\frac{|\kappa|\cdot d^3}{4} + O(d^4)$, see brown region in Figure \ref{fig:blob2}.
\end{theorem}
This theorem is proved in the Appendix (Section~\ref{sec:area}). It allows to accurately estimate curvature 
$\kappa_n$ at contour point $n$ using triple cliques $c_i(p)=(\pmi,p,\ppi)$ of orientation $\theta_{i(n)}$ 
consistent with the tangent at point $n$. Assuming  $\kappa_n\geq0$, Theorem~\ref{th:area} implies that 
the number of triple cliques $c_i(p)$ in a vicinity of point $n$ with configuration $(0,1,0)$ is
$$A(\kappa_n,d_i)\approx\frac{|\kappa_n|\cdot d_i^3}{4},$$ 
which is the brown area in Figure \ref{fig:blob2}. In case $\kappa_n\leq0$, the same number $A$ 
corresponds to configurations $(1,0,1)$.

Assuming that triple cliques $c_i(p)$ of configurations $(0,1,0)$ or $(1,0,1)$ have penalties $w_i$
depending on orientation $i$, the overall cost of all such cliques is
$$\sum_i \sum_p   w_i\cdot \delta(X_{c_i(p)}) \;\; = \;\; \sum_i \sum_{n: i(n)=i} w_i\cdot A(\kappa_n,d_i)$$
where
\begin{equation} \label{eq:delta}
 \delta(X_c):=  \begin{cases} 1\;\;\;\;\;\;\mbox{if}\;\; X_c=(1,0,1)\;\mbox{or}\;(0,1,0) \\ 0 \;\;\;\;\;\mbox{otherwise} \end{cases}
\end{equation}
is an indicator function. Choosing penalty
\begin{equation}\label{eq:weights}
w_i = \frac{4\cdot\Delta\theta_i}{d_i^3}
\end{equation}
converts the right hand side in the equation above into
$$\sum_i \sum_{n: i(n)=i} \frac{4\cdot\Delta\theta_i}{d_i^3} \cdot A(\kappa_n,d_i)\;\; = \;\; \sum_{n=1}^N 
 |\kappa_n|\cdot\Delta\theta_{i(n)}.$$
In combination with \eqref{eq:contK_partial} and \eqref{eq:contK_approx} this demonstrates that
\begin{equation} \label{eq:energy}
\int_C \kappa^2 \cdot ds  \;\; \approx \;\;    \sum_i \sum_p   w_i \cdot \delta(X_{c_i(p)})
\end{equation}
which is our main technical result concluding this section.

\begin{figure}
\includegraphics[width =\linewidth]{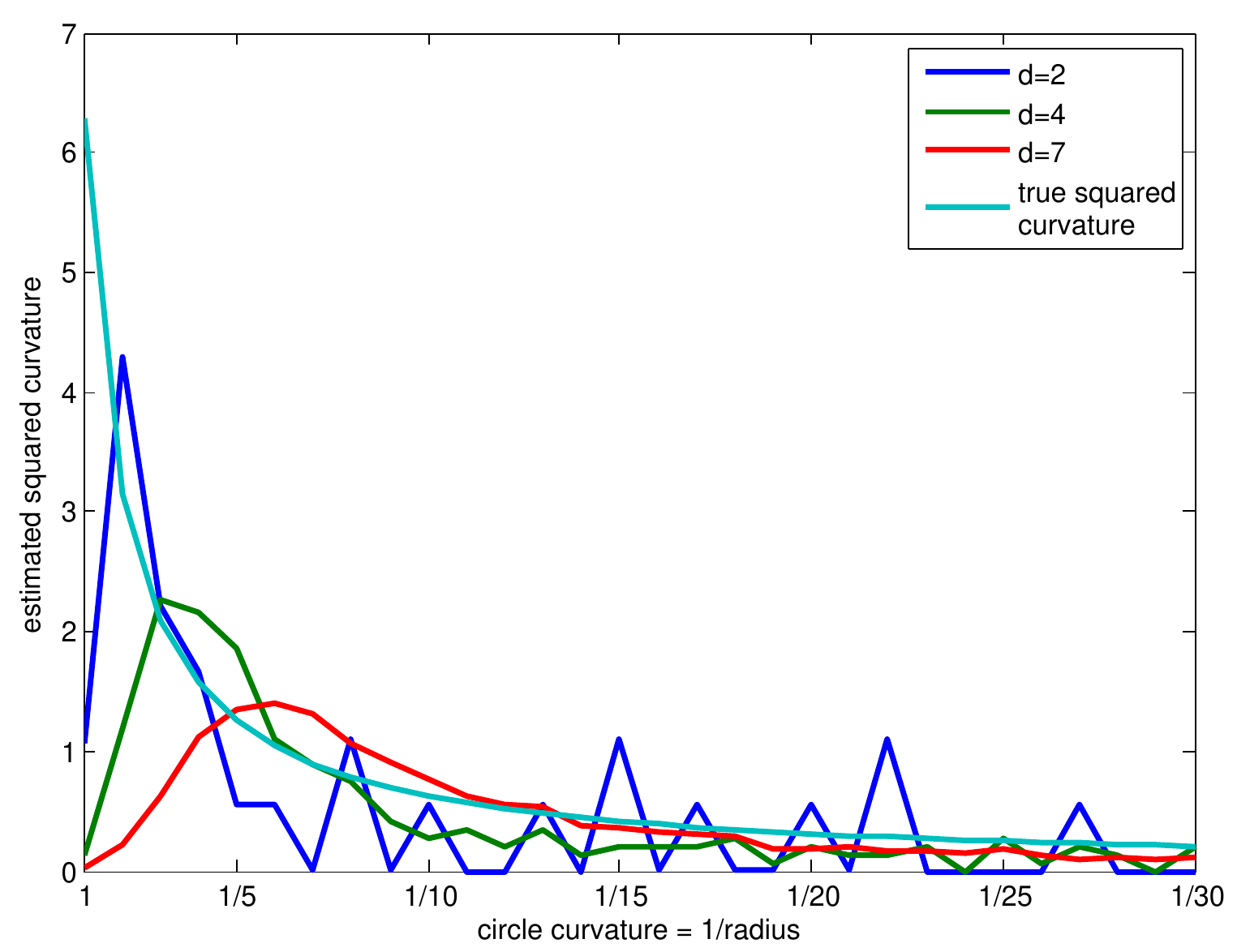}
\caption{Accuracy of our squared curvature model. The proposed estimated squared curvature (vertical axis) is computed on a circle of decreasing total curvature (i.e. increasing radius, horizontal axis) based on increasing clique size $d$. To approximate large curvature small clique sizes $d$ are required, whereas large clique sizes are more accurate to measure large curvature and yield less noisy measurements. For curvature values $\kappa > \frac{1}{d}$ we underestimate curvature, since in these cases the circle fits inside the neighborhood. This  limits the estimated curvature energy by the number of pixels in the circle, i.e. its area, which grows quadratically. }  
\label{fig:circle_experiment}
\end{figure}

\subsection{Accuracy of Our Curvature Model}

\begin{figure}
\tabcolsep0.5mm
\begin{tabular}{ccc}
\includegraphics[width = 0.33\linewidth]{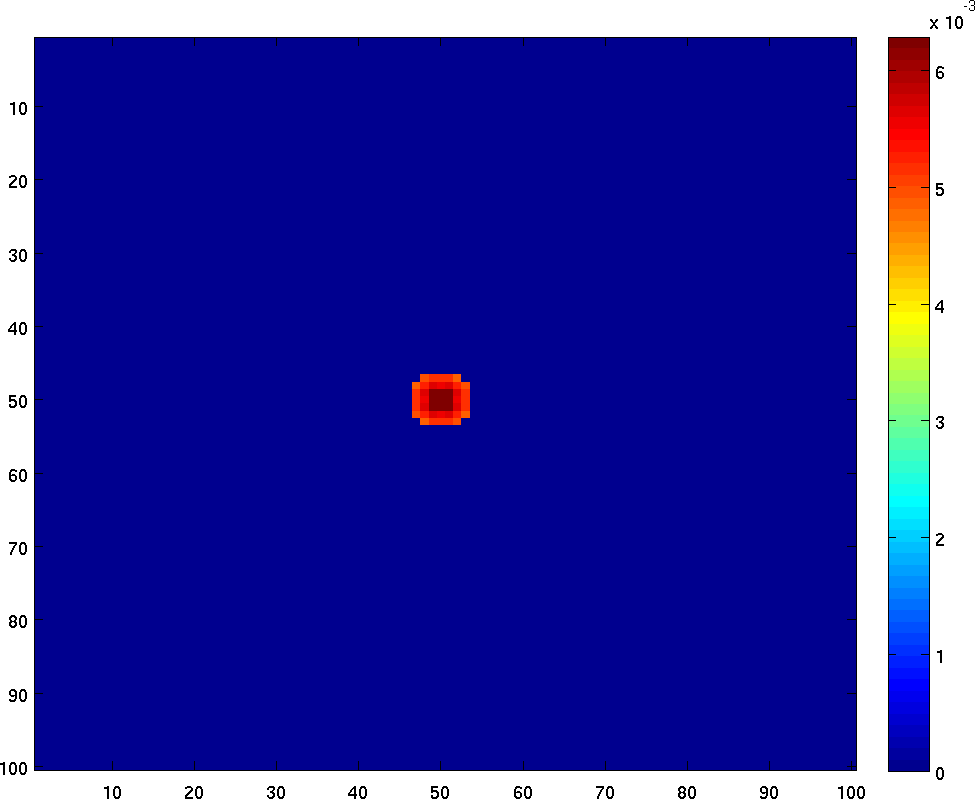}&
\includegraphics[width = 0.33\linewidth]{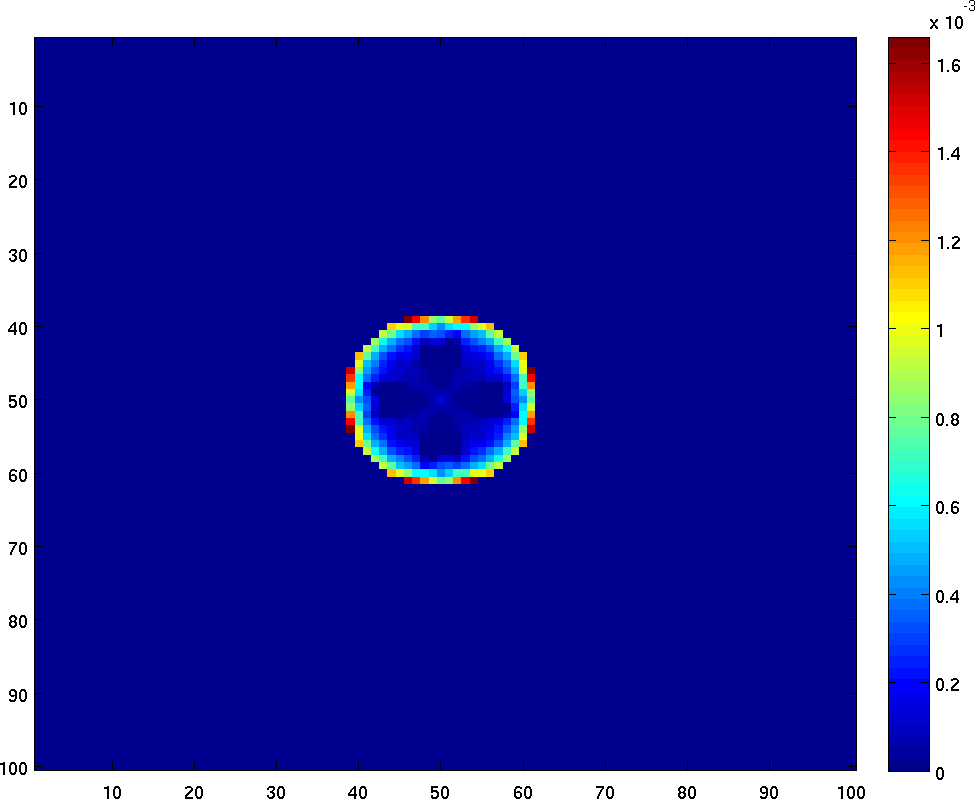}&
\includegraphics[width = 0.33\linewidth]{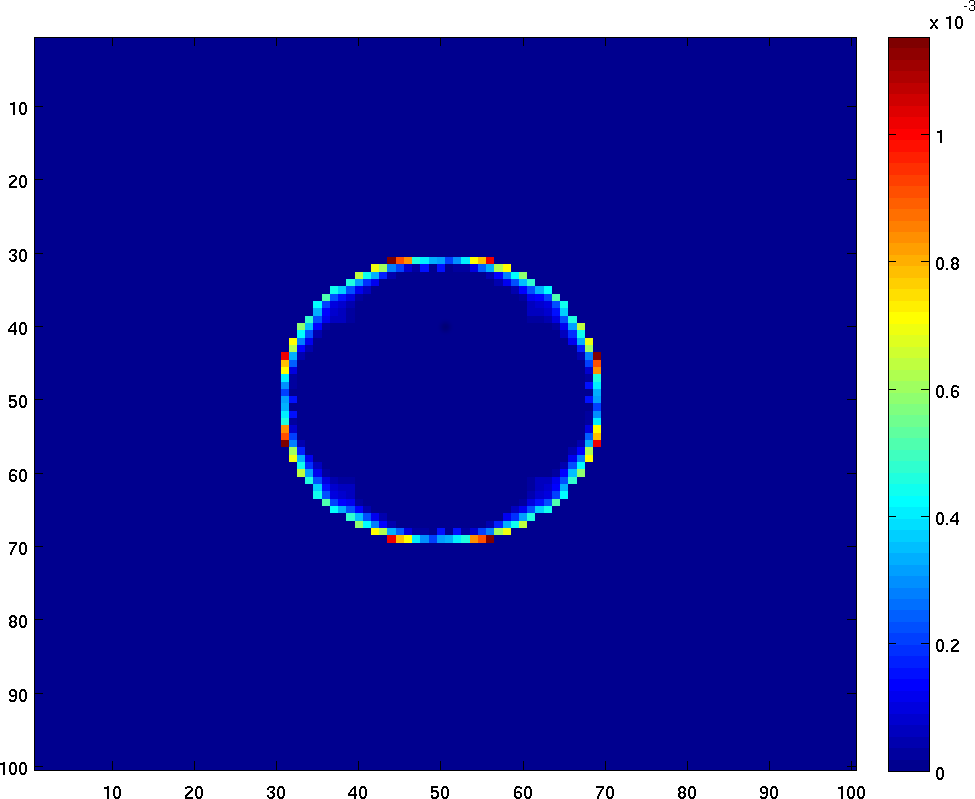}\\
(a) $\kappa = 1/4$ & (b) $\kappa = 1/12$ & (c) $\kappa =1/20$  
\end{tabular}
\caption{Response images showing for each pixel of a circle of increasing radius (i.e. decreasing squared curvature) its contribution to the curvature energy, see Figure \ref{fig:circles}. The computations are based on a clique size $d = 9$. The images show that the brown area in Figure \ref{fig:blob2} is approximated with our approach. (a) also shows that large curvature is underestimated for large values of $d$ (compare Figure~\ref{fig:circle_experiment}).}
\label{fig:responses}
\end{figure}

To demonstrate the accuracy of the proposed squared curvature model we
generated a sequence of circles with decreasing total squared curvature
(increasing radius $r$). We compared the computed curvature energy
\eqref{eq:energy} obtained with different neighborhood sizes to the
correct integral of squared curvature of the circle, $2\pi
\kappa$. The plots in Figure \ref{fig:circle_experiment} show that we
indeed approximate squared curvature. We can also conclude that for
smaller neighborhoods we more accurately measure larger curvature
values but obtain more noisy results. In contrast, our results are
less noisy for larger neighborhood masks but yield less accurate
measurements for larger curvature. This is because our measurement is
limited by the area of the circle, which is very small for small
radius (large curvature) and grows quadratically. These findings are
supported by Figure \ref{fig:responses}, which shows the contribution
of each pixel to the measured curvature energy. To alleviate the issue
of underestimating curvature for larger values of $d$ the image can be
scaled to subpixel accuracy according to the increase in neighborhood
size or stronger regularization can be used, see
Section~\ref{sec:triplecliquelength} in the experiments.  
% Figure
% \ref{fig:largeneighborhoods} illustrates this effect.  Alternatively,
% stronger regularization can be used to obtain smooth results without
% increasing the image resolution, however at the cost of losing
% details.

% This subsection can groups all synthetic experiments 
% that we did to evaluate the accuracy of our curvature model
% (Figures 9-10). This is instead of section 5.3 (experiemnts of circles).
% I think it is OK to have this before details on our clique optimization
% as these experiments do not require any optimization. 

\section{Optimization}

In \eqref{eq:energy} we derived our new squared curvature model, which can be used as a regularization energy
\begin{equation}\label{eq:regularizer}
	E_{curv}(X) = \sum_i \sum_p  w_i \cdot \delta(x_p,x_{\ppi},x_{\pmi}).
\end{equation} 
Let $a,b,c$ denote the three pixels corresponding to a triple clique
$c_i(p)$ given in \eqref{eq:tripleclique}. The function $\delta$ defined in
\eqref{eq:delta} assigns each triple clique the value 1 if its
configuration is (1,0,1) or (0,1,0). The corresponding triple clique
energy in \eqref{eq:regularizer} naturally decomposes into a unary term, two pairwise submodular terms and
a pairwise supermodular term without additional auxiliary variables
\begin{align}
\delta(x_a,x_b,x_c) &= x_a (1-x_b)x_c + (1-x_a)x_b(1-x_c) \nonumber\\
&= x_b + x_ax_c-x_ax_b-x_bx_c. 
\end{align}

% Hence, we obtain the final curvature energy
% \begin{equation}\label{eq:finalenergy}
% 	E_{curv}(X) = \sum_i \sum_p  w_i (x_p + x_{\pmi}x_{\ppi}-x_{\pmi}x_p-x_px_{\ppi})
% \end{equation}
 
%Interestingly, our clique potentials naturally reduce from  triplets to pairwise and unary terms without addition of auxiliary variables. 
In general, energies of lower order  are considered to be easier to optimize. This is an additional benefit of our integral geometry model that we get for 'free'. 

Our energy is nevertheless non-submodular and, therefore, cannot be optimized globally.
We use a new method called Local Submodular Approximations with Trust Region (LSA-TR) proposed in our companion paper \cite{LSATR:companion}. The method is efficient and obtains state-of-the-art results that are very accurate on a wide range of applications, which we show in our companion paper.

We compare our results in terms of accuracy and runtime
to other optimization approaches (TRW-S \cite{kolmogorov-pami-2006}, LBP \cite{pearl-1982}, QPBO-I \cite{rother-et-al-cvpr-2007} - an extension of QPBO \cite{boros-et-al-1991}) in the experimental section \ref{sec:opteff}. Throughout all experiments we use the same set of default parameters
for LSA-TR.

\begin{figure}
\centering
\includegraphics[height=0.4\linewidth]{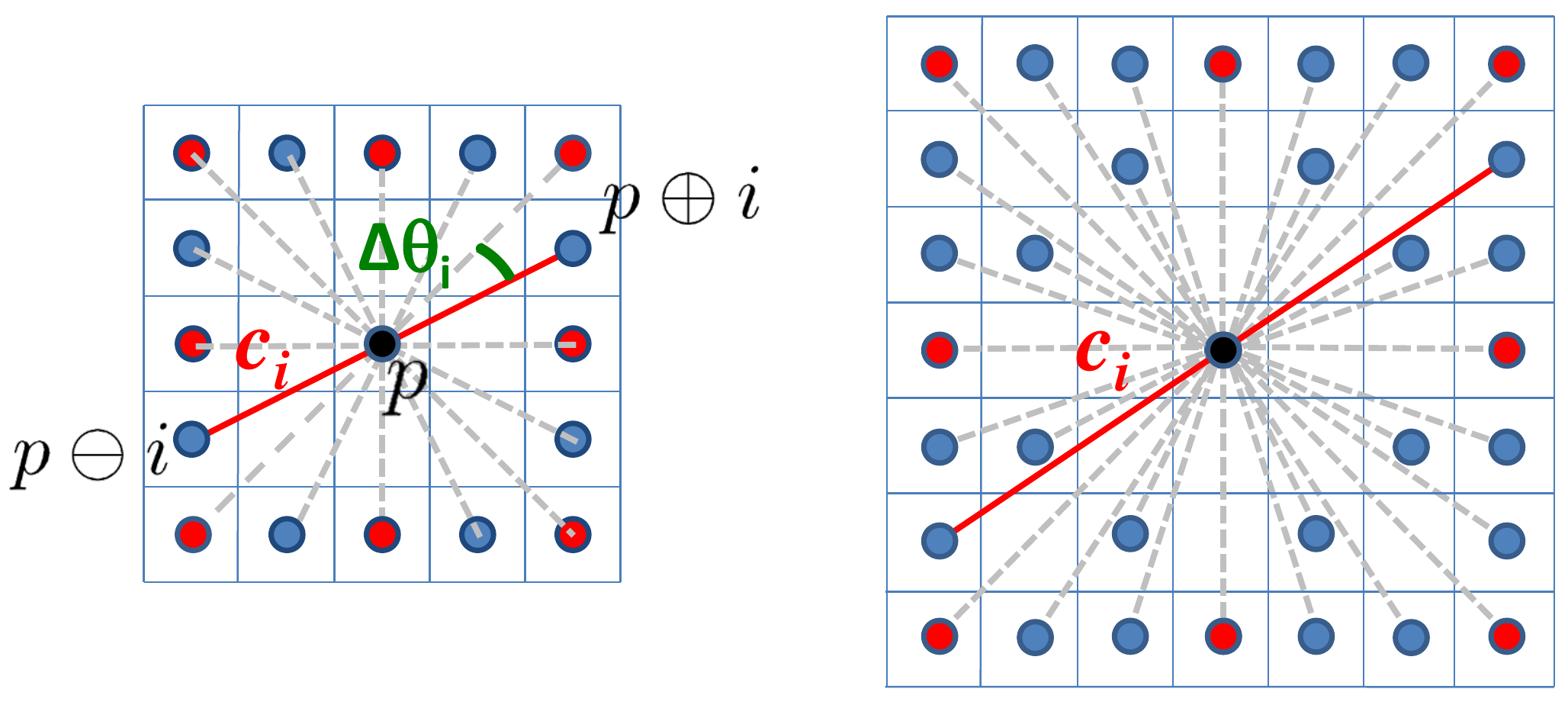}
\caption{Defining neighborhood systems of different sizes. For each $d$ we define a box of size $(2d+1)\times(2d+1)$. For each orientation in this neighborhood we then push the neighbor as far towards the boundary of the box as possible (see the red points, which are different from standard neighborhoods) to obtain the approximate distance $d$ of each neighbor to the central clique point. Triplets are then composed of opposite neighbors and the central point of the neighborhood. On the left we show an example for $d=2$ corresponding to a $5\times5$ neighborhood, on the right for $d=3$ corresponding to a $7\times7$ neighborhood.}
\label{fig:nhexample}
\end{figure}

{\bf Neighborhood Definition}
It remains to construct the neighborhood for defining our triple cliques. We use a fixed box of size $(2d+1)\times(2d+1)$ for constructing triple cliques of size $d$ for a set of orientations $\Theta$ around the central pixel (we denote the specific case of only horizontal and vertical neighbors in a $3\times3$ neighborhood by ``$2\times2$'' throughout the experimental section). Neighbors are moved as far towards the edge of the box as possible, see Figure~\ref{fig:nhexample}. The angle difference $\delta_i$ between neighboring triple cliques can then be computed as the angle between subsequent cliques in this neighborhood system. The corresponding weights $w_i$ are then given by \eqref{eq:weights}.

\section{Experiments}
This section evaluates our squared curvature approach 
and compares our results to previous methods for curvature regularization. 

We show results for binary segmentation and inpainting which  minimize the following energy
\begin{align}
\label{eq:segenergy}
	E(X) = \sum_i D(x_i, I_i) + \lambda E_{curv}(X).
\end{align}
The data term $D(x_i, I_i)$ depends on the pixel color of the i-th pixel in the image $I$. The weight $\lambda$ balances the impact of the regularizer with respect to the appearance term.
We use a Gaussian of variance 0.4 for modeling the foreground and background data term in our experiments. The mean values are 0 and 0.6 for foreground and background respectively for the camera man examples and 0 and 1 for the Don Quixote image. 
For inpainting, we use the same energy but set the data term to 1 for foreground and background simultaneously for the region to be inpainted.

%\subsection{Subpixel Accuracy}
\subsection{Triple Clique Length $d$} \label{sec:triplecliquelength}
\begin{figure}
\tabcolsep0.5mm
\centering
\begin{tabular}{cc}
\includegraphics[width = 0.39\linewidth]{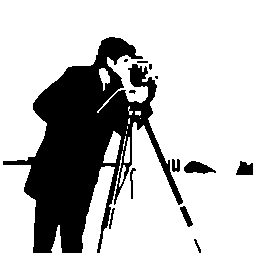} &
\includegraphics[width = 0.39\linewidth]{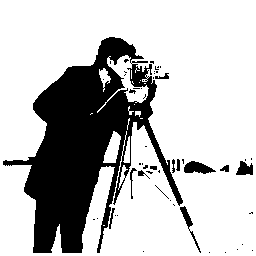} \\
(a) $3\times3$, $\lambda = 0.1$, 1x & (b) $5\times5$, $\lambda = 0.1$, 1x  \\ 
\includegraphics[width = 0.39\linewidth]{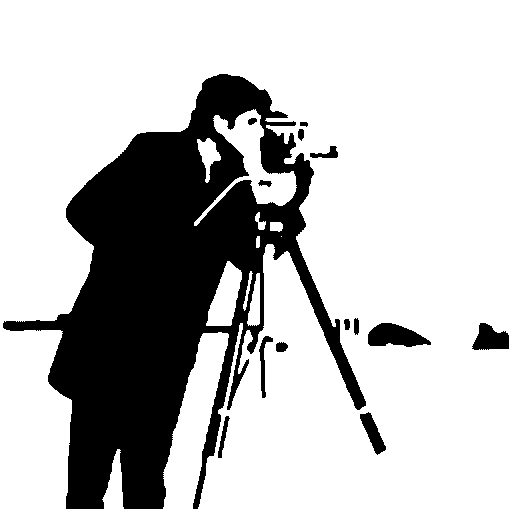} &
\includegraphics[width = 0.39\linewidth]{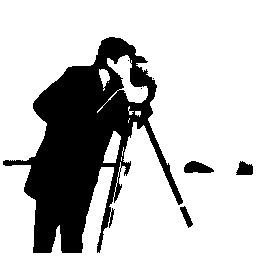} \\
(c) $5\times5$, $\lambda = 0.8$, 3x & (d) $5\times5$, $\lambda = 0.8$, 1x
\end{tabular}
\caption{Segmentation results on the cameraman image showing the effect of using large triple cliques. (a) Segmentation result with novel squared curvature regularizer using a $3\times3$ neighborhood (8-connectivity) ($\lambda = 0.1$). (b) The same segmentation using a $5\times5$ neighborhood (16-connectivity). Note the noise which is due to underestimating large curvature locally (see text for details). (c) By scaling the image (and adapting $\lambda$ accordingly), we avoid underestimation while taking advantage of a bigger angular resolution. (d) Similar effects but with less details can be achieved by increasing the regularizer weight instead of scaling the image.}
\label{fig:largeneighborhoods}
\end{figure}

We first show the effect of larger triple cliques in Figure~\ref{fig:largeneighborhoods}. 
The larger the length of the triple cliques the more we underestimate large curvature values (see Figure~\ref{fig:circle_experiment}) leading to noise in the segmentation. 
Yet, we would like to use large neighborhoods to attain a larger angular resolution of our clique orientations as shown in the next section. As a remedy we propose to scale the image according to triple clique length $d$ in order to achieve subpixel accuracy. In this way we do not underestimate large curvature and avoid angular artifacts. %in order to achieve subpixel accuracy.
Alternatively, one can increase regularization to reduce the noise but this
might lead to less details in the segmentation. These points are
illustrated in Figure~\ref{fig:largeneighborhoods}.
% The first image shows the segmentation result using a small 3x3
% neighborhood (amounting to 4 different triplet orientations per
% pixel). Applying the same algorithm with a larger neighborhood (5x5,
% equivalent to 8 orientations) yields a rather noisy result
% (\textit{center} of Figure~\ref{fig:largeneighborhoods}, compare
% e.g. the region under the tripot and around the camera).  By
% increasing the image size to obtain subpixel accuracy we obtain a
% clean segmentation result. 

%\subsection{Increasing Angular Resolution and Curvature Weight}

\subsection{Angular Resolution}
\begin{figure}
\tabcolsep0.3mm
\begin{tabular}{ccc}
\includegraphics[width = 0.33\linewidth]{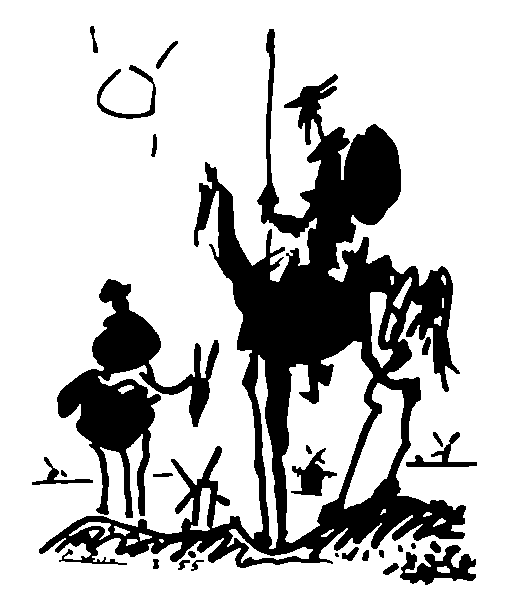}&
\includegraphics[width = 0.33\linewidth]{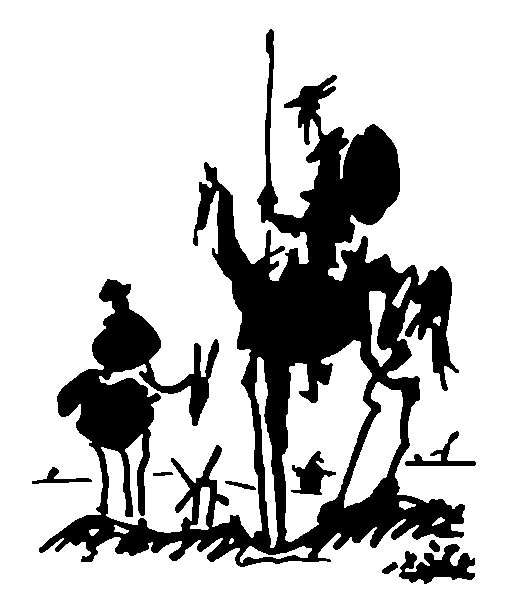}&
\includegraphics[width = 0.33\linewidth]{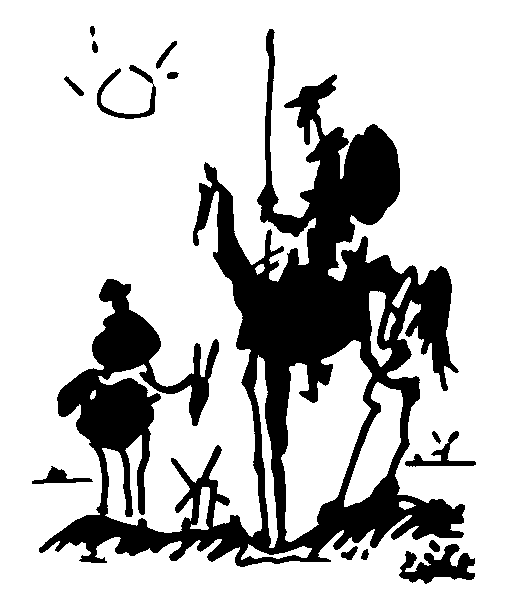}\\
$2\times2$, small & $3\times3$, small & $7\times7$, small \\
\includegraphics[width = 0.33\linewidth]{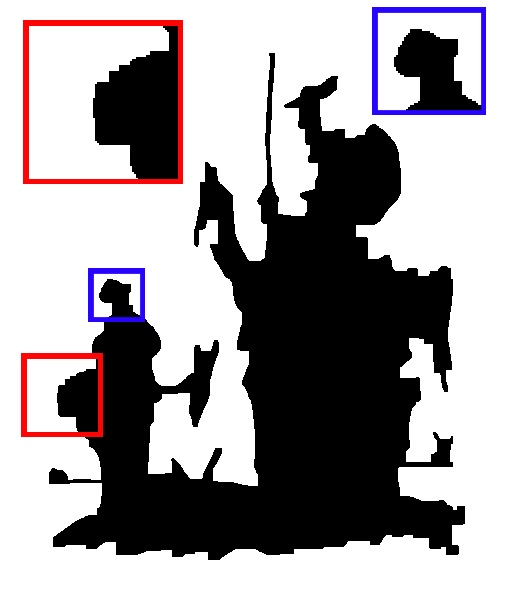} &
\includegraphics[width = 0.33\linewidth]{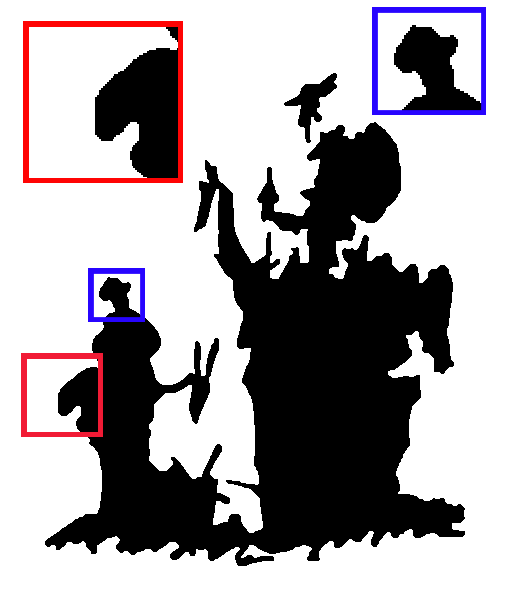} &
\includegraphics[width = 0.33\linewidth]{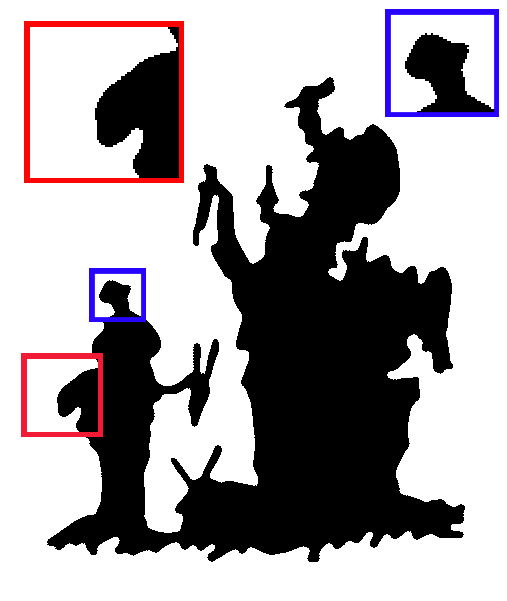} \\
 $2\times2$, large & $3\times3$, large & $7\times7$, large  
\end{tabular}
\caption{Segmentation results for Picasso's ink drawing for small and large curvature regularization weight and increasing neighborhood size. Note that the choice of the neighborhood size only becomes important for larger regularization weights. The images clearly show how 'blocky' structures disappear with larger neighborhood sizes.}
\label{fig:angularres}
\end{figure}

We now show the effect of increasing angular resolution on the segmentation results. Figure~\ref{fig:angularres} shows results for angular resolutions $\theta$ of 90 degress ($2\times2$ neighborhood), 45 degrees ($3\times3$) and 12.5 degrees ($7\times7$). For small curvature weight the appearance is strong yielding comparable results for all angular resolutions. For very large curvature weight block structures become apparent. This is due to the fact that the algorithm is blind to curvature for points whose tangent direction is not in our clique set. Accordingly, for only 90 degree resolution horizontal and vertical edges become dominant in the result. For 45 degree resolution diagonal edges appear as well. For larger resolution we obtain smooth boundaries. 

%\begin{figure*}
%\begin{tabular}{ccc}
%\includegraphics[width = 0.24\linewidth]{images/blockiness/3x3_4nh_0_9.png} &
%\includegraphics[width = 0.24\linewidth]{images/blockiness/3x3_8nh_0_9.png} &
%\includegraphics[width = 0.24\linewidth]{images/blockiness/5x5_16nh_8.png} \\
%3x3, 4nh & 3x3, 8nh & 5x5, 16nh
%\end{tabular}
%\caption{Blockiness}
%\label{fig:largeneighborhoods}
%\end{figure*}

\subsection{Inpainting Results}
\begin{figure}
 \begin{tabular}{ccc}
 \includegraphics[width = 0.26\linewidth]{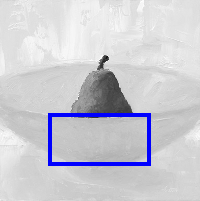} &
 \includegraphics[width = 0.26\linewidth]{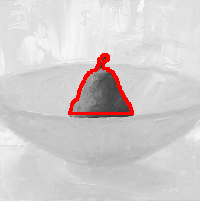} &
 \includegraphics[width = 0.26\linewidth]{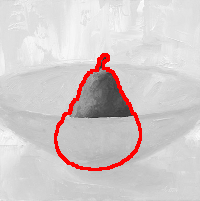} \\
 \includegraphics[width = 0.26\linewidth]{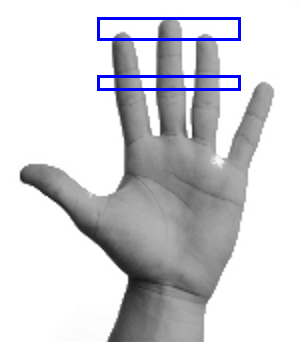} &
 \includegraphics[width = 0.26\linewidth]{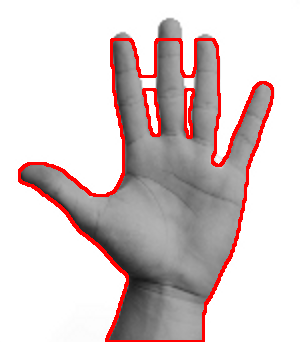} &
 \includegraphics[width = 0.26\linewidth]{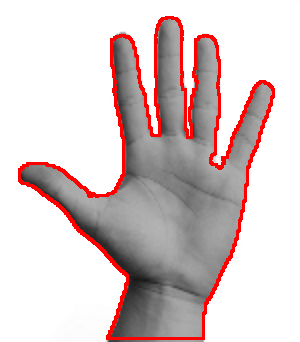} \\
 a) Input with mask & b) Length & c) Curvature 
 \end{tabular} 
 \caption{Results for binary inpainting with the proposed curvature regularizer compared to length based inpainting. a) Original image with masked regions to be inpainted, b) Result of length based inpainting, c) Result of curvature based inpainting with the novel approach.}
 \label{fig:inpainting}
 \end{figure}

In this section we show results for the task of binary inpainting
demonstrating that our approach is able to reconstruct shape with
minimal curvature instead of length. We show two examples in Figure
\ref{fig:inpainting}, for which we mask larger portions of the
image for which we ``occlude'' its data: the data term in these regions is set to a constant value for
both foreground and background. For the remaining part of the image we
use a data term based on a Gaussian intensity model with fixed mean
and variance for foreground and background. Figure
\ref{fig:inpainting} shows the results of our experiments and a
comparison to length based inpainting. The results clearly show that
we minimize the curvature of the object boundary, while length finds
the shortest connection between boundary segments.

\subsection{Optimization and Efficiency} \label{sec:opteff}

\begin{figure}
\centering
\includegraphics[width=0.8\linewidth]{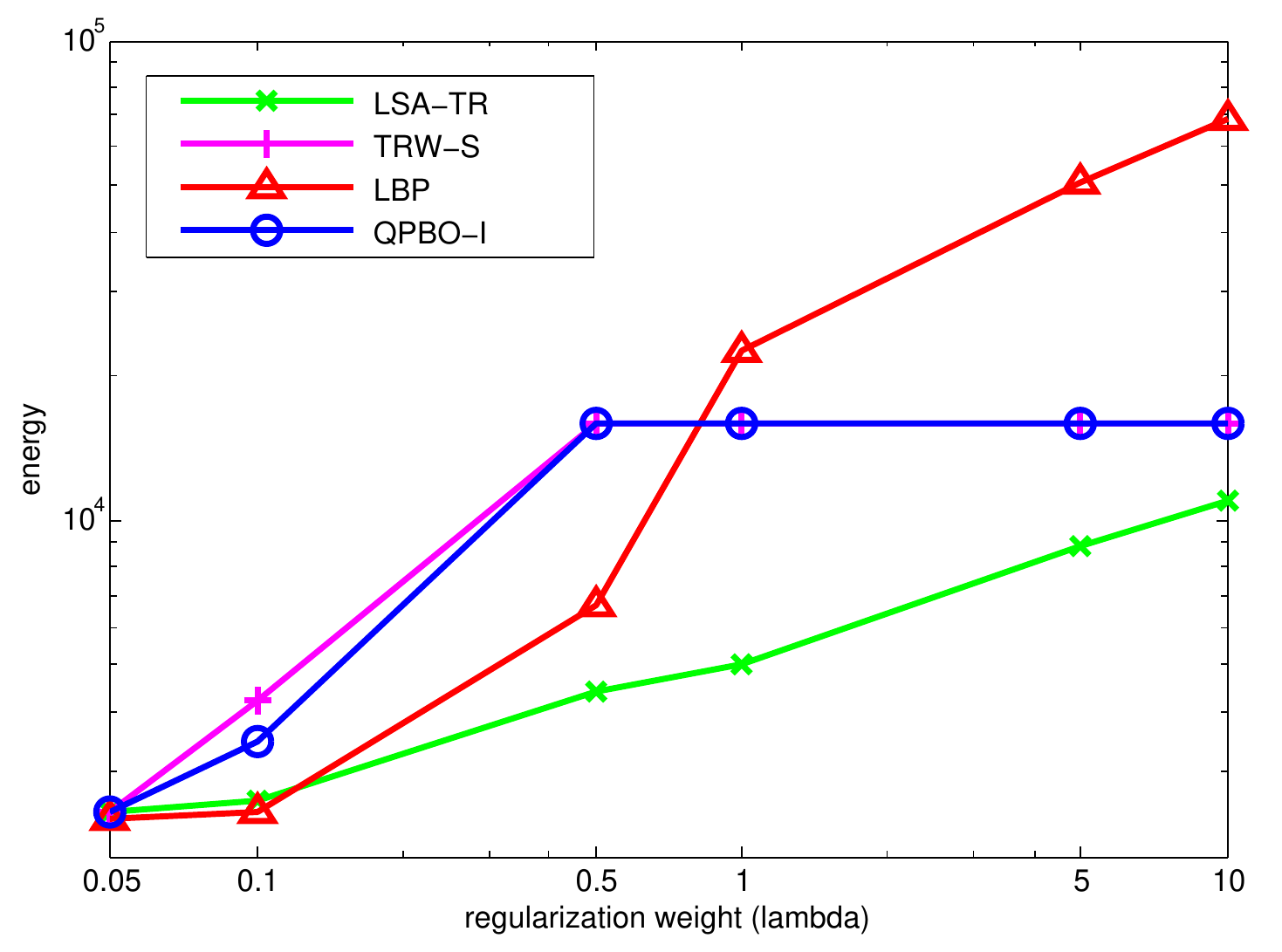}
\caption{Comparison of optimized energies based on different regularization strengths $\lambda$ for the following optimization algorithms: LSA-TR \cite{LSATR:companion}, TRW-S \cite{kolmogorov-pami-2006}, LBP \cite{pearl-1982}, QPBO-I \cite{rother-et-al-cvpr-2007}. LSA-TR clearly outperforms the other algorithms yielding reasonable results even for strong regularization, while the other approaches break down with trivial (TRWS, QPBO-I) or extremely noisy solutions (LBP).} 
\label{fig:energy_plot}
\end{figure}

We evaluated the performance and runtime of different optimization strategies for minimizing the energy in \eqref{eq:segenergy}. 
In particular, we compared QPBO-I \cite{rother-et-al-cvpr-2007}, TRW-S \cite{kolmogorov-pami-2006}, Loopy Belief Propagation (LBP) \cite{pearl-1982} and LSA-TR \cite{LSATR:companion}. For TRW-S we stopped computations after 50,000 steps without convergence of the algorithm. Running both TRW-S and LBP even for 500,000 steps did not improve the results. Figure~\ref{fig:energy_plot} shows the energies we obtain for the respective methods on the vertical axis plotted against the regularization weight $\lambda$ on the horizontal axis. 
Of all the tested algorithms, LSA-TR finds the lowest energy for our problem for almost all values $\lambda$. This is especially true for higher curvature weights, where QPBO-I and TRW-S compute trivial solutions of higher energy with almost all pixels labeled as background. In particular, QPBO is unable to label any pixels at all for $\lambda \geq 0.5$ and therefore QPBO-I cannot improve the result either. The high energies we get with LBP for $\lambda \geq 1$ correlates with the extremely noisy results the algorithm returns.
Figure~\ref{fig:runtimes} shows a comparison in runtime revealing that TRW-S is least efficient of the four algorithms, whereas LSA-TR is the fastest optimization method. 
%In sum we notice that FTR is the only practically applicable optimization method for computing segmentations regularized with the new model for squared curvature in terms of runtime and quality.
The proposed method also compares remarkably well to other approaches that compute squared curvature. For the results in Figure~\ref{fig:comparison} the runtimes are as follows: Heber et al. \cite{heber-et-al-eccv-2012} 1 to 5 minutes, Schoenemann et al. \cite{schoenemann-etal-ijcv-2012} 10 minutes to 3.5 hours and only El-Zehiry and Grady's method is fast with 10 seconds per image. Strandmark and Kahl's approach \cite{strandmark-kahl-emmcvpr-2011} is only slightly faster than Schoenemann's.

\begin{figure}
\small           
\begin{center}
\begin{tabular}{ccccccc} \hline 
$\lambda$ & QPBO & TRW-S & LBP & LSA-TR & (7x7)\\ \hline
0.05 & 0.7 (2\%) & 751 & 680 & 0.8 & 0.7\\
0.1 & 9.3 (30\%) & 764 & 682 & 1.4 & 1.6 \\
0.5 & 15.2 (100\%) & 715 & 709 & 3.9 & 9.2 \\
1 & 15.4 (100\%) & 713 & 795 & 10.6 & 48.2 \\
5 & 15.3 (100\%) & 714 & 794 & 16.5 & 88 \\
10 & 15.4 (100\%) & 703 & 797 & 7.9 & 110 \\\hline 
\end{tabular} 
 							
\caption{Runtime comparison for different optimization methods ($3\times3$ neighborhood). For QPBO \cite{boros-et-al-1991} the percentage of unlabeled pixels is indicated in parenthesis. For TRW-S we stopped computations after 50,000 steps without convergence. LSA-TR is also listed for a $7\times7$ neighborhood with $\lambda\times 10$ to obtain comparable results.}
\label{fig:runtimes} 
\end{center}  
\end{figure}

\subsection{Comparison to Other Approaches}
\begin{figure*}
\tabcolsep0.2mm
\begin{tabular}{ccccccc}
\includegraphics[width = 0.14\linewidth]{donquixote.jpg} &
\includegraphics[width = 0.14\linewidth]{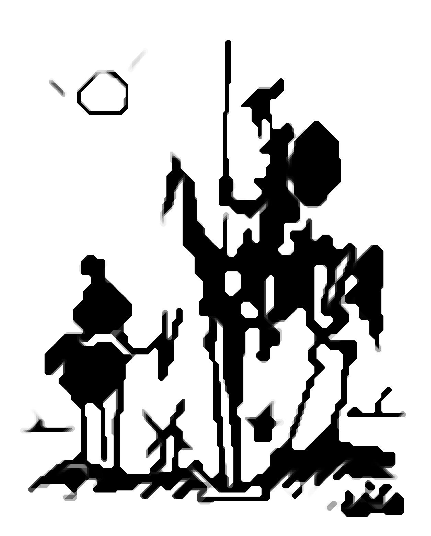} &
\includegraphics[width = 0.14\linewidth]{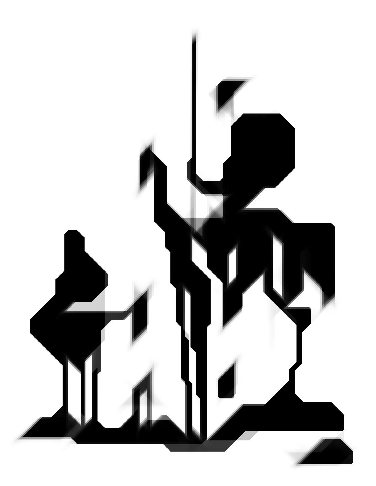} &
\includegraphics[width = 0.14\linewidth]{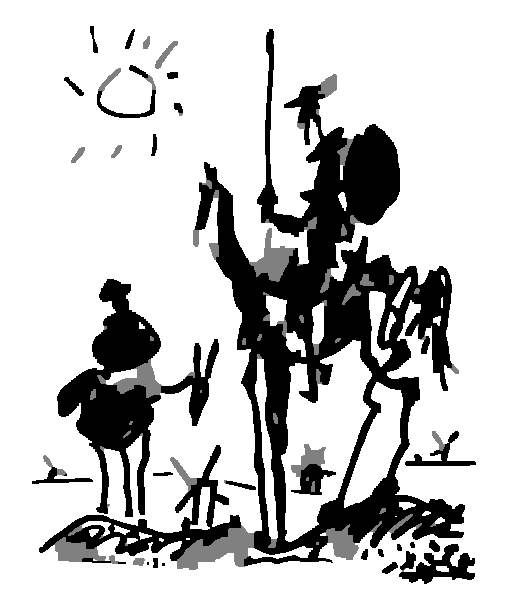} &
\includegraphics[width = 0.14\linewidth]{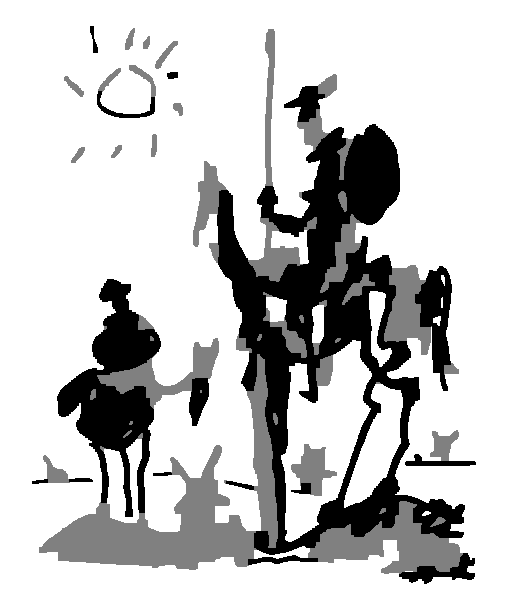} &
\includegraphics[width = 0.14\linewidth]{int_7x7.png} &
\includegraphics[width = 0.14\linewidth]{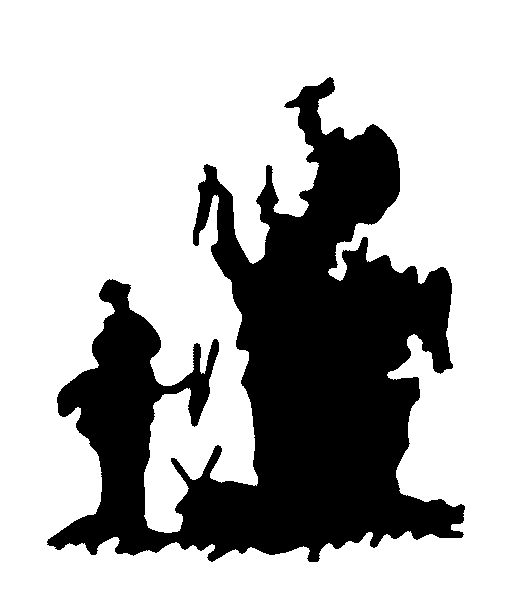} \\
(a) original & (b) \cite{heber-et-al-eccv-2012} small & (c) \cite{heber-et-al-eccv-2012} large & (d) \cite{elzehiry-grady-cvpr-2010} small & (e) \cite{elzehiry-grady-cvpr-2010} large & (f) ours small & (g) ours large \\
\includegraphics[width = 0.14\linewidth]{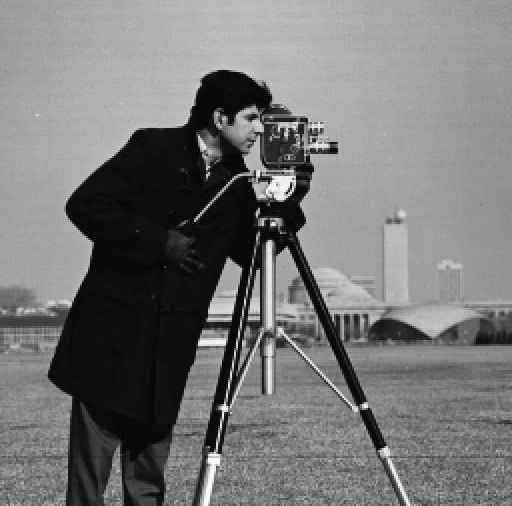} &
\includegraphics[width = 0.14\linewidth]{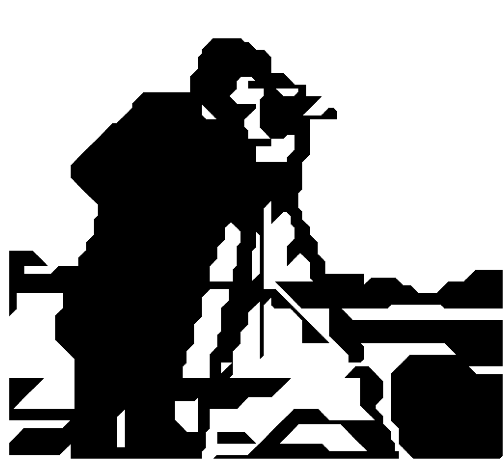} &
\includegraphics[width = 0.14\linewidth]{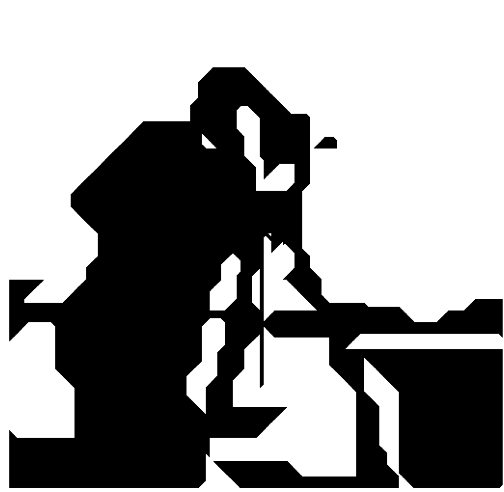} &
\includegraphics[width = 0.14\linewidth]{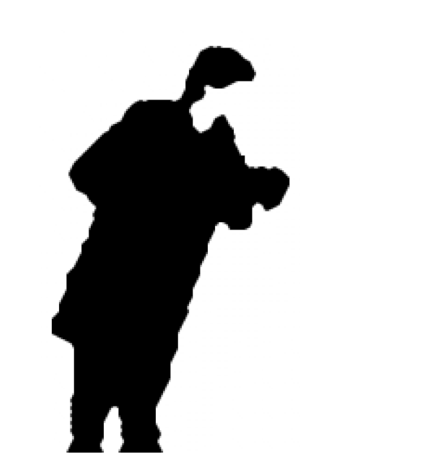} &
\includegraphics[width = 0.14\linewidth,height=70pt]{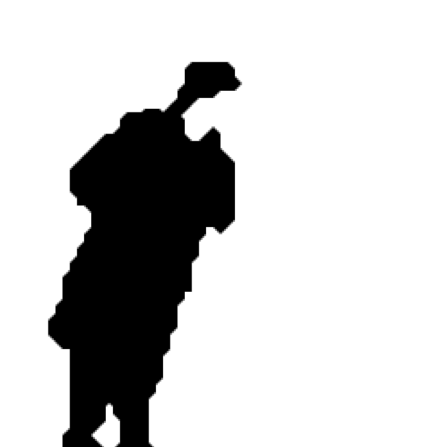} &
\includegraphics[width = 0.14\linewidth]{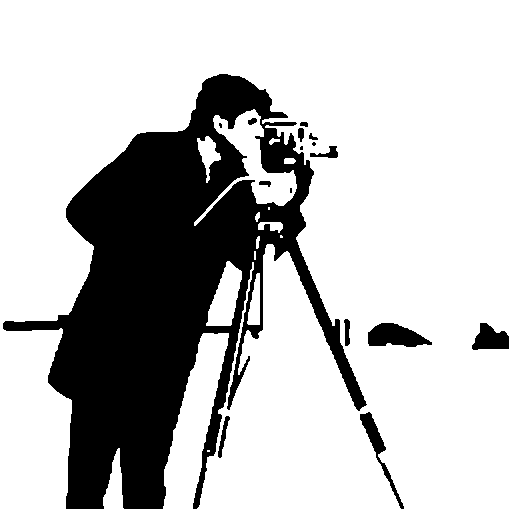}  & 
\includegraphics[width = 0.14\linewidth]{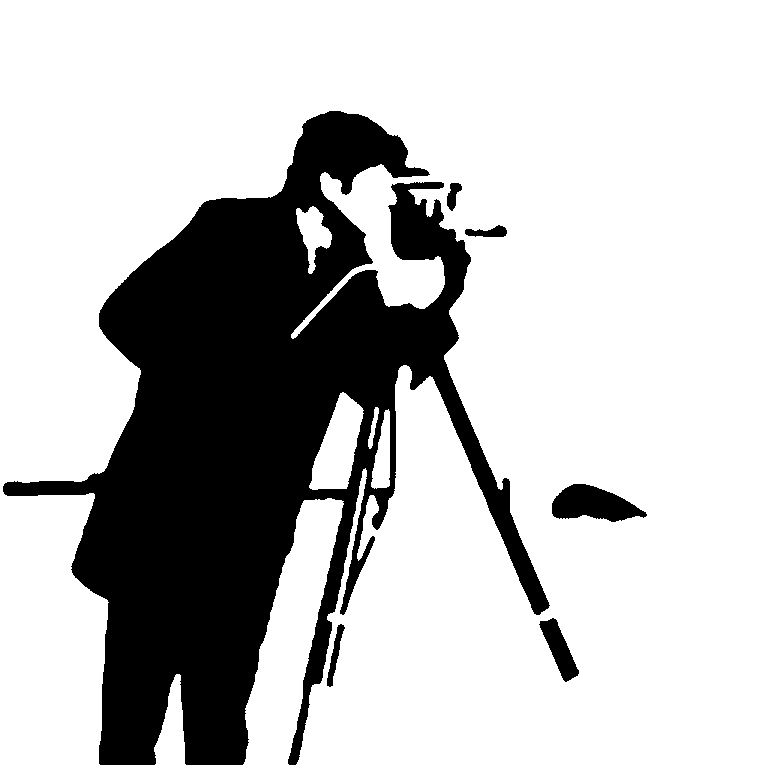} \\
(h) original & (i) \cite{schoenemann-etal-ijcv-2012} small & (j) \cite{schoenemann-etal-ijcv-2012} large & (k) \cite{strandmark-kahl-emmcvpr-2011} small & (l) \cite{strandmark-kahl-emmcvpr-2011} large  & (m) ours small  & (n) ours  large  \\
\end{tabular}
\caption{Comparison of our results for 7x7 neighborhood to previous curvature regularizers for smaller and larger curvature weight, (b+c) elastica by Heber et al. \cite{heber-et-al-eccv-2012}, (d+e) El-Zehiry and Grady \cite{elzehiry-grady-cvpr-2010}, (i+j)Schoenemann et al. \cite{schoenemann-etal-ijcv-2012}, (k+l) Strandmark and Kahl \cite{strandmark-kahl-emmcvpr-2011}. Grey pixels are unassigned, which is due to QPBO leaving pixels unlabeled in El-Zehiry and Grady \cite{elzehiry-grady-cvpr-2010} - using QPBO-I does not improve the results.}
\label{fig:comparison}
\end{figure*}

We compare our curvature regularizer to previous approaches in Figure
\ref{fig:comparison} where we show results for small and large regularization
for the approach by Heber et al. \cite{heber-et-al-eccv-2012}, Schoenemann et al. \cite{schoenemann-etal-ijcv-2012}, Strandmark and Kahl \cite{strandmark-kahl-emmcvpr-2011}
and El-Zehiry and Grady \cite{elzehiry-grady-cvpr-2010}. 
For all of these methods strong
artifacts are evident, which become worse for stronger
regularization. Heber et al. \cite{heber-et-al-eccv-2012} compute elastica (i.e. length and squared curvature regularization), which preserves long structures, but also introduces block artifacts. El-Zehiry and Grady are limited by 90 degree resolution, i.e. their results are composed of blocks and contain large regions of unlabeled pixels QPBO \cite{boros-et-al-1991} did not label. Schoenemann et al. \cite{schoenemann-etal-ijcv-2012} obtain block artifacts with increasing regularization. Strandmark and Kahl \cite{strandmark-kahl-emmcvpr-2011} do not preserve fine details and also suffer from some angular resolution artifacts. In contrast, our method (neighborhood size $7\times7$) preserves fine details in the segmentation and produces clean object boundaries for weak and strong regularization without requiring excessive runtimes.

% The reason for this lies in the nature of our
% approximation: boundary parts with large local curvature fall within
% the span of the triplets used to estimate the squared
% curvature. Figure~\ref{fig:circle_experiment} shows that for those
% regions the curvature is badly approximated. Instead, for curvature
% above some critical value, the measure suddently drops proportionally
% to the area of the measured region.

% \begin{figure*}
% \tabcolsep0.5mm
% \begin{tabular}{cccc}
% \includegraphics[width = 0.24\linewidth]{images/donquixote.jpg}&
% \includegraphics[width = 0.24\linewidth]{images/don_5x5_7_2.png} &
% \includegraphics[width = 0.24\linewidth]{images/don_5x5_7_2.png} &
% \includegraphics[width = 0.24\linewidth]{images/don_5x5_7_2.png} \\
% \includegraphics[width = 0.24\linewidth]{images/cameraman.jpg}&
% \includegraphics[width = 0.24\linewidth]{images/cameraman_3x3_0_1_x1.png} &
% \includegraphics[width = 0.24\linewidth]{images/cameraman_5x5_0_8_x2.png} &
% \includegraphics[width = 0.24\linewidth]{images/cameraman_7x7_2_7_x3.png} \\
% Original & 3x3 & 5x5 & 7x7 
% \end{tabular}
% \caption{Segmentation result of Pablo Picasso's Don Quixote ink drawing based on color information and minimal squared curvature using a novel curvature regularization method.}
% \label{fig:teaser}
% \end{figure*}

\section{Conclusion and Future Work} \label{sec:conclusions}
In this paper we proposed a novel approach to squared curvature computation and regularization. We gave an integral geometric derivation of our method which justifies that straight triple cliques can be used to measure squared curvature. Our triple cliques decompose into a set of submodular and supermodular pairwise cliques, which can be efficiently optimized by LSA-TR. We showed that our approach works for high angular resolutions and thus does not suffer from grid artifacts as do previous methods. The results demonstrate that we outperform these methods in terms of quality and efficiency. \\
This approach naturally extends to higher dimensions for the regularization of 3D surfaces, which will be important e.g. for 3D reconstruction approaches. We will leave this for future work.

\section{Appendix: Proof of Theorem 1} \label{sec:area}
In this appendix we will prove Theorem \ref{th:area}. 
% To measure the squared curvature of the contour of an object we assume that its curvature is constant within the vicinity of a boundary point. At each contour point the contour is then approximated by a circle of radius $r$. The key idea of our approach is to measure a weighted partial circle area by means of triplets (line segments of fixed length $d$). 
%
We will use an integral geometric argument to show that squared curvature is related to the brown area in Figure~\ref{fig:blob2}.

\begin{figure}[b!]
\begin{tabular}{cc} 
\includegraphics[width = 0.45\linewidth]{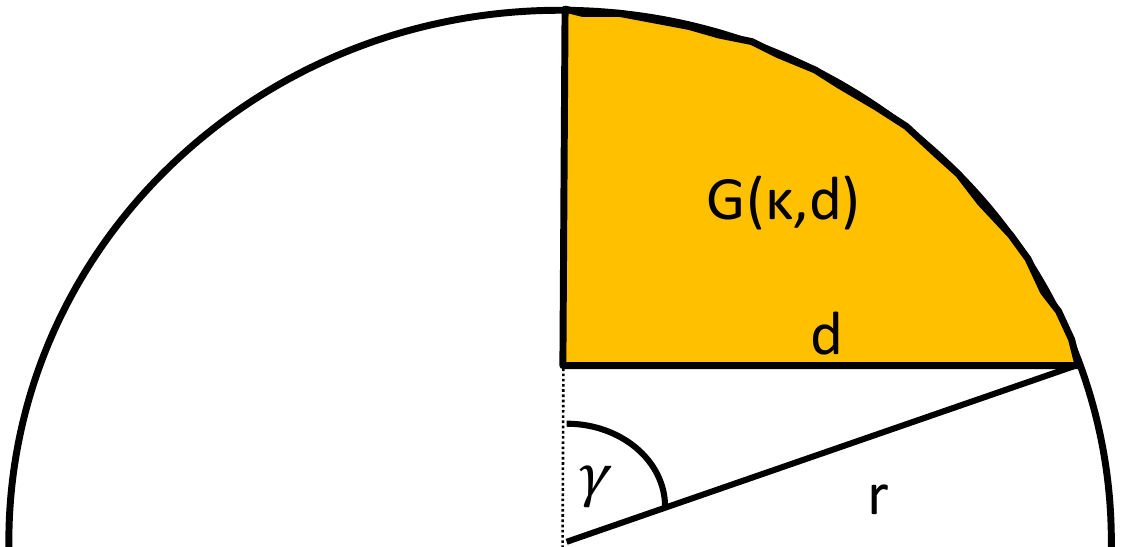} &
\includegraphics[width = 0.45\linewidth]{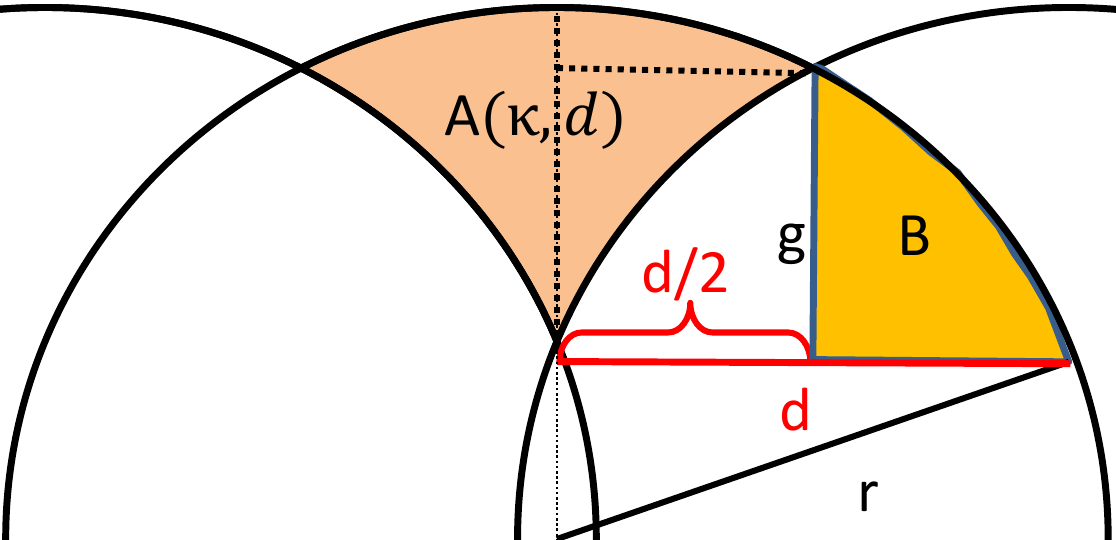} \\
(a) & (b) \\
\end{tabular}
\caption{(a) Computing the partial area $G(\kappa, d)$ of a circle above a half chord of length $d$, (b) Computing the shaded area $A(\kappa,d)$}
\label{fig:circles}
\end{figure}

We first state that the partial area of the circle of curvature $\kappa = \frac{1}{r}$ above a half chord of length $d < r$ in Figure~\ref{fig:circles}(a) is given by 
\begin{eqnarray}
G(\kappa,d)  &:=& \frac{1}{2 \kappa^2} \gamma - \frac{d}{2 \kappa} \cos{\gamma}  \nonumber\\
&=& \frac{1}{2 \kappa^2} \arcsin \kappa d - \frac{d}{2} \sqrt{\frac{1}{\kappa^2}-d^2}.
\end{eqnarray}
Based on this expression we can now derive the area $A(\kappa,d)$ in Figure~\ref{fig:circles}(b) with respect to a half chord of length $d$ in a circle of radius $r$
\begin{eqnarray} 
A(\kappa,d) \!\!\!\!\!&:=&  \!\!\!\! 2\Bigg[G\left(d,\kappa\right) - 2\underbrace{\left(G\left(d,\kappa\right) - G\left(\frac{d}{2},\kappa\right) - \frac{gd}{2}\right)}_{area B}\Bigg] \nonumber\\
&=& \!\!\!\! -\frac{1}{\kappa^2} \arcsin{\kappa d} + 2\frac{1}{\kappa^2} \arcsin \frac{\kappa d}{2} \nonumber\\ 
&& \!\!\!\! - d \sqrt{\frac{1}{\kappa^2}-d^2} + d\sqrt{\frac{1}{\kappa^2}-\frac{d^2}{4}}  
\end{eqnarray}
Using Taylor approximation w.r.t. $d$ we obtain
\begin{equation} \label{eq:taylor}
A(\kappa,d) \approx \frac{d^3 \kappa}{4}.
\end{equation}
% Integrating This expression over rotation angle $\alpha$ (i.e. we rotate the chord d over all possible angles) we finally approximate the area in Figure~\ref{fig:?}. This area is directly related to the integral of the squared curvature $\kappa^2$ along the contour of the object
% \begin{equation}\label{eq:weights}
% \int_\alpha \frac{d^3}{32r} d \alpha = \int_{\partial S} \frac{d^3}{32 r^2} ds = \int_{\partial S} \frac{d^3}{32} \kappa^2 ds.
% \end{equation}

{\small
\bibliographystyle{ieee}
\bibliography{newlib}

\begin{thebibliography}{10}\itemsep=-1pt

\bibitem{boros-et-al-1991}
E.~Boros, P.~Hammer, and X.~Sun.
\newblock Network flows and minimization of quadratic pseudo-boolean functions.
\newblock {\em Technical report RRR 17-1991, RUTCOR}, 1991.

\bibitem{boykov-jolly-iccv-2001}
Y.~Boykov and M.-P. Jolly.
\newblock Interactive graph cuts for optimal boundary \& region segmentation of
  objects in n-d images.
\newblock In {\em Int. Conf. of Computer Vision}, Vancover, Canada, 2001.

\bibitem{BK:iccv03}
Y.~Boykov and V.~Kolmogorov.
\newblock Computing geodesics and minimal surfaces via graph cuts.
\newblock In {\em International Conference on Computer Vision (ICCV)}, 2003.

\bibitem{Bruckstein01}
A.~Bruckstein, A.~Netravali, and T.~Richardson.
\newblock Epi-convergence of discrete elastica.
\newblock {\em Applicable Analysis}, 79(1-2):137--171, 2001.

\bibitem{elzehiry-grady-cvpr-2010}
N.~El-Zehiry and L.~Grady.
\newblock Fast global optimization of curvature.
\newblock In {\em Conf. Computer Vision and Pattern Recognition}, 2010.

\bibitem{LSATR:companion}
L.~Gorelick, Y.~Boykov, O.~Veksler, I.~BenAyed, and A.~Delong.
\newblock Submodularization for {Q}uadratic {P}seudo-{B}oolean {O}ptimization.
\newblock In {\em ArXiv}, 2013.

\bibitem{heber-et-al-eccv-2012}
S.~Heber, R.~Ranftl, and T.~Pock.
\newblock Approximate envelope minimization for curvature regularity.
\newblock In {\em Proceedings of the 12th international conference on Computer
  Vision - Volume Part III}, European Conf.\ on Computer Vision, pages
  283--292, 2012.

\bibitem{kolmogorov-pami-2006}
V.~Kolmogorov.
\newblock Convergent tree-reweighted message passing for energy minimization.
\newblock {\em IEEE Transanctions on Pattern Analysis and Machine.
  Intelligence.}, 28:1568--1583, October 2006.

\bibitem{pearl-1982}
J.~Pearl.
\newblock Reverend bayes on inference engines: A distributed hierarchical
  approach.
\newblock In {\em National Conference on Artificial Intelligence}, pages
  133--136, 1982.

\bibitem{rother-et-al-cvpr-2007}
C.~Rother, V.~Kolmogorov, V.~Lempitsky, and M.~Szummer.
\newblock Optimizing binary mrfs via extended roof duality.
\newblock In {\em Conf.\ Computer Vision and Pattern Recognition}, pages 1--8,
  2007.

\bibitem{schoenemann-etal-ijcv-2012}
T.~Schoenemann, F.~Kahl, S.~Masnou, and D.~Cremers.
\newblock A linear framework for region-based image segmentation and inpainting
  involving curvature penalization.
\newblock {\em Int. Journal of Computer Vision}, 2012.

\bibitem{shekhovtsov-etal-dagm-2012}
A.~Shekhovtsov, P.~Kohli, and C.~Rother.
\newblock Curvature prior for {MRF}-based segmentation and shape inpaint.
\newblock In {\em DAGM}, 2012.

\bibitem{strandmark-kahl-emmcvpr-2011}
P.~Strandmark and F.~Kahl.
\newblock Curvature regularization for curves and surfaces in a global
  optimization framework.
\newblock In {\em EMMCVPR}, pages 205--218, 2011.

\end{thebibliography}
}

\end{document}